\pdfoutput=1

\documentclass[11pt]{article}

\usepackage{acl}

\usepackage{times}
\usepackage{latexsym}
\usepackage{graphicx}
\usepackage{booktabs}

\usepackage{setspace}
\usepackage[vlined]{algorithm2e}
\SetAlCapNameFnt{\fontsize{10pt}{12pt}\selectfont}
\SetAlCapFnt{\fontsize{10pt}{12pt}\selectfont\normalfont}

\usepackage[frozencache]{minted}
\usepackage[utf8]{inputenc}

\DeclareUnicodeCharacter{2014}{\dash}
\DeclareRobustCommand\dash{%
  \unskip\nobreak\thinspace\textemdash\allowbreak\thinspace\ignorespaces}

\bgroup 
  \catcode`\_=\active
  
\egroup
\usepackage[T1]{fontenc}

\usepackage{microtype}

\usepackage{xspace,bm}
\makeatletter
\newcommand*{\ifbold}{%
  \ifx\f@series\my@test@b
    \expandafter\@firstoftwo
  \else
    \expandafter\@secondoftwo
  \fi
}
\newcommand*{\my@test@b}{b}
\makeatother
\newcommand\Adaptor{Adapt%
  \ifbold
    {\scalebox{0.65}{$\bm{\mathcal{O}}$}r}
    {\scalebox{0.65}{$\mathcal{O}$}\kern0.1pt\relax r}
\xspace}
\title{\Adaptor: Objective-Centric Adaptation Framework for Language Models}

\author{Michal Štefánik$^{1,2}$ \and Vít Novotný$^1$ \and Nikola Groverová$^2$ \and Petr Sojka$^1$ \\
  $^1$Faculty of Informatics, Masaryk University, Czech Republic \\
  $^2$Gauss Algorithmic \\
\vspace*{-2\baselineskip}}

\begin{document}
\maketitle
\begin{abstract}
Progress in natural language processing research is catalyzed by the possibilities given by the widespread software frameworks.
This paper introduces the \Adaptor library\footnote{\url{github.com/gaussalgo/adaptor}} that transposes the traditional model-centric approach composed of pre-training + fine-tuning steps to objective-centric approach, composing the training process by \textit{applications} of selected \textit{objectives}.
We survey research directions that can benefit from enhanced objective-centric experimentation in multi-task training, custom objectives development, dynamic training curricula, or domain adaptation.
\Adaptor aims to ease the reproducibility of these research directions in practice. Finally, we demonstrate the practical applicability of \Adaptor in selected unsupervised domain adaptation scenarios.
\end{abstract}

\vspace*{-.5\baselineskip}
\begin{flushright}
\mbox{\llap{“}The\,measure\,of\,intelligence\,is\,the\,ability\,to\,change.\rlap{”}}\\
― Albert Einstein
\end{flushright} 
\vspace*{-\baselineskip}

\section{Introduction}

Recent development in Natural Language Processing (NLP) heavily benefits from a high level of maturity of open-source frameworks, such as Fairseq \citep{ott2019fairseq} or HuggingFace Transformers \citep{Wolf2019HuggingFacesTS}. Thanks to the standardized interfaces, these libraries allow for immediate experimentation with the most recent research results, practically fostering the speed of further progress in the area. 
While their use is seamless for countless conventional use-cases of transformer models and fine-tuning to a specific end-task \citep{devlin-etal-2019-bert,Radford2018gpt}, divergence from this framework requires feasible, but elaborate and complex customizations, increasing the risk of logical errors and complicating the reproducibility of experiments. A characteristic group of problems requiring significant changes to the standard pipeline are multi-step and multi-task adaptations.
\looseness=-1

This paper introduces the \Adaptor library, which aims to simplify the more complex training processes that their training objectives can easier describe. \Adaptor challenges the conventional \textit{model-centric} framework, where data and task selection are constrained by the requirements of the selected language model architecture.
Instead, it introduces an \textit{objective-centric} training pipeline, with Objective as the central abstraction of the process.

\newcommand\modul[1]{\textsf{\small#1}}
\begin{figure}
\includegraphics[scale=1.33]{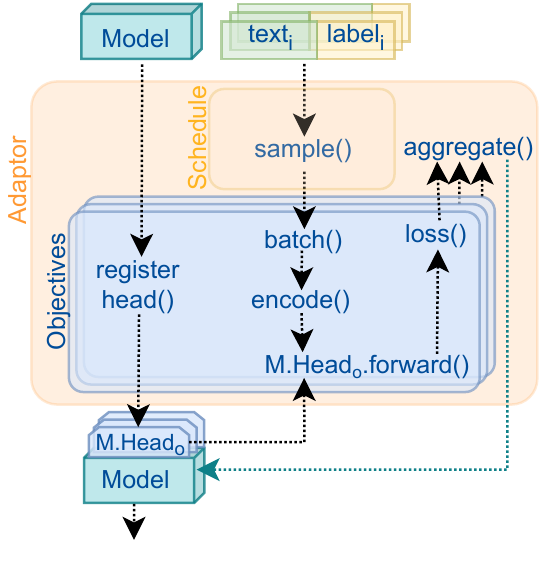}%
\vspace{-5mm}
    \caption{Overview of \Adaptor's objective-centric training framework: \modul{Objective} 1)~registers its compatible head on top of the shared model, 2)~performs specific input encoding, and 3)~compute loss value based on its output. A~\modul{Schedule} implements a specific sampling curricula and \Adaptor aggregates and propagates objectives' losses and performs optimization.}
    \label{fig:correlations}
\end{figure}

The \Adaptor framework aims to help NLP researchers and practicioners engage in projects that include any of the following:
\begin{itemize}
    \item \textbf{Multi-objective training}: when training a language model on more than one task or data set, including languages, \Adaptor can significantly simplify the custom code base that needs to be implemented. Even if the objective is custom, the user can avoid adjustments to other parts of the training pipeline.
    \item \textbf{Custom data schedule}: when users need to perform dynamic data sampling, \Adaptor allows them to implement a custom Schedule (see Figure~\ref{fig:schedule}), leaving the data and model adjustment logic intact. This simplifies systematic experimentation and reproducibility, and minimizes the risk of errors.
    \item \textbf{Objectives design \& evaluation}: \Adaptor exposes top-level declaration of training objectives, which enables easy experimentation with custom objectives. Objective-level monitoring can provide custom behavioural insights and allows for pruning less promising experiments earlier in the lengthy training process, saving computational costs.
    \item \textbf{Robustness evaluation}: The objective-centric paradigm provides an easy robustness estimation by evaluating on out-of-distribution samples.
    In the standard \textit{sequential} adaptation scenario, objective-centric evaluation exposes characteristic flaws of adaptation, like exposure bias or catastrophic forgetting.
\end{itemize}

This paper is structured as follows: Section~\ref{sec:background} provides an overview of recent work demonstrating the potential of multi-objective training in domain and task adaptation.
Section~\ref{sec:adapt_frameworks} also describes other software frameworks applicable for similar use cases.
Section~\ref{sec:framework_design} describes the design of \Adaptor, showing the users how to confidently integrate novel objectives and schedules. 
In Section~\ref{sec:experiments}, we describe and implement a set of non-trivial, yet promising domain adaptation experiments using \Adaptor and collect their results.
As \Adaptor remains under active development, we close in Section~\ref{sec:conclusion} with an outline of the upcoming features.
We welcome contributions of novel objectives and schedules.

\section{Background}
\label{sec:background}

This section provides an overview of recent work that demonstrates the potential of multi-objective training and schedules that motivated the design of \Adaptor. Our overview consists of a non-exhaustive list of applications that \Adaptor aims to make more accessible for practical use and in future research.

\subsection{Multi-Task Training}
\label{sec:multitask_training}

Multi-task training has a long history in both traditional machine learning \cite{caruana97_machine_learning} and in deep learning \cite{Crawshaw2020MultiTaskLW}.
This section describes examples of multi-task (i.e.\ multi-objective) training, outlining its benefits and potential.

Under some circumstances, multi-task training enhances distributional robustness of neural models. \citet{multitask_eliminates_biases} demonstrate this on adversarial data sets, exposing common heuristic biases of the language models~\cite{McCoy2019RightFT}. 
Enhanced model generalization can also be achieved by introducing one or more latent tasks that do not directly correspond to the end task but reflect specific desired properties of the model. One of a few studies in this direction is Sharpness-Aware Minimisation of \citet{Foret2021SharpnessAwareMF}, performing multi-objective training on image classification using cross-entropy and a novel, sharpness-aware objective, reflecting the model's monotonicity on the local neighborhood.
In context of Neural Machine Translation (NMT), \citet{wang-sennrich-2020-exposure} incorporate Minimum Risk Training (MRT) objective \cite{Ranzato2016SequenceLT}, optimising an arbitrary sequence-level measure of outputs.
In composition with the traditional token-level cross-entropy objective, MRT improves distributional robustness.
 
By aggregating multiple objectives, \citet{xie2019unsupervised} show that combining sentence classification objective with maximizing representation consistency to augmented samples fosters data efficiency.

The intuition on the benefits of multi-task training presumes that by optimizing the training by multiple cost functions, the final model is less prone to the weaknesses of a specific task~\cite{Collobert:2011:NLP:2078183.2078186}, possibly reflecting on higher-level, task-invariant properties of language~\cite{Bengio2013RepresentationLA}.

\subsection{Data-Sampling Schedules}
\label{sec:data_sampling}

Exposing a model to training samples in a systematic schedule, also referred to as a \textit{curriculum}, can lead to an improvement of the accuracy of the final model~\cite{Bengio2009CurriculumL}. 
While the positive effects of more complex schedules based on sample ``difficulty'' with transformers remain to be explored, multiple studies show the potential of confidence-based sampling to improve accuracy and generalization. 
Biased samples can be identified, according to model's confidence \cite{Pleiss0EW20, swayamdipta-etal-2020-dataset} or using Bayesian methods such as the Product of Experts~\cite{Hinton2002TrainingPO}. 
Then, they can be either eliminated \cite{LeBras2020AdversarialFO} or downweighted \cite{Utama2020TowardsDN}.

More complex scheduling methods are applied in training NMT models.
\citet{NIPS2015_e995f98d} use decay schedule to sample from both references and the previous outputs of a NMT model, minimizing the discrepancy between training and inference.
\citet{zhang-etal-2019-bridging} successfully use the same sampling strategy in a sequence-level objective.
The results of \citet{lu-etal-2020-mixed} underline the potential of sampling in NMT training, suggesting that the accuracy of transformers on reported MT benchmarks can be outperformed by simpler RNN models by combining objectives in decay schedule.

Despite the reported improvements, we find that custom scheduling strategies are rarely used.
We attribute this to their complicated integration into the standard training process.
To foster the research and applicability of scheduling methods, \Adaptor makes the implementation of custom scheduling strategies easy, comprehensible, and reproducible.

\subsection{Domain Adaptation}
\label{sec:adapt_strategies}

Objective-centric frameworks are well-suited for domain adaptation techniques, where \Adaptor provides support for combining traditional end-task objectives with unsupervised adaptation or auxiliary-task objectives in a user-selected schedule.
The goal of domain adaptation is to maximize performance on a specific data domain, often denoted as the \textit{adapted} or \textit{target domain} \cite{Saunders2021DomainAA}.

Perhaps the most common adaptation approach using pre-trained language models is to continue pre-training on unsupervised samples of the adapted domain \cite{luong-manning-2015-stanford,10.1093/bioinformatics/btz682,Beltagy2019SciBERT}.
This approach has been successfully extended in various directions.
For instance, \citet{dontstoppretraining2020} show that adapting to a shared task on different domain can enhance accuracy of the eventual application.
If supervised data is sparse, other auxiliary tasks, described earlier in Section~\ref{sec:multitask_training}, can be used as concurrent objectives~\cite{xie2019unsupervised}. 

In cases where larger volumes of data of given task is available in a different language, adaptation using cross-lingual transfer can be considered.
Pre-trained language models show that cross-lingual transfer works well with large-data unsupervised objectives \cite{lample2019cross}, but it can also be applied for low-resource supervised objective, such as very low-resource translation~\cite{neubig-hu-2018-rapid}.

If even unsupervised target-domain data is sparse, another option is to subset arbitrary unsupervised sources to automatically identify samples of adapted domain, by applying domain classifier \cite{jiang-zhai-2007-instance,ElSahar2019ToAO}.
If the boundary between the training and the adapted domain is known, an auxiliary objective can minimise a discrepancy of representations between the training and possibly low-resource target domain~\cite{Chadha2018ImprovingAD}.

Despite the possibilities, adaptation can also introduce undesired biases. In the scope of NMT, adaptation can cause problems of ``catastrophic forgetting'', when the model experiences performance degradation on the originally well-performing domains \cite{Saunders2021DomainAA}, or ``exposure bias'', when the model overfits the non-representative specifics of the target domain, such as the artifacts of data collection \cite{Ranzato2016SequenceLT}.
Additionally, by normalizing a single type of bias, such as lexical overlap \cite{McCoy2019RightFT}, the model might degrade its accuracy on other domains~\cite{Utama2020TowardsDN}.
Addressing multiple biases concurrently~\cite{2020arXiv201003338W} can mitigate this problem.

\Adaptor allows the knowledgeable user to construct a reproducible and robust adaptation pipeline using native multi-objective evaluation. Covering multiple domains in separate objectives, \Adaptor can expose the above pitfalls, without the need to implement complex separate evaluation routines.

\begin{figure*}
\scalebox{0.8}{
  \begin{minipage}{\linewidth}
  \inputminted[linenos]{python}{soubor-fig01.py}
  \end{minipage}
}
\caption{\Adaptor provides a convenient base for implementing custom sampling schedules. ParallelSchedule in the figure demonstrates an implementation of the schedule sampling the update objectives in rotation. Further, the sampling can be easily conditioned on the state of Objectives such as the recent outputs, loss, or metrics evaluations.}
\label{fig:schedule}
\end{figure*}

\subsection{Related Software Frameworks}
\label{sec:adapt_frameworks}

The \textit{Adapters} architecture \cite{pmlr-v97-houlsby19a}, having only a small set of parameters, might be a good fit when performing adaptation of transformer with modest hardware or data. 
Recently, the AdapterHub library \cite{pfeiffer-etal-2020-adapterhub} makes training and sharing of Adapters convenient.
Compared to \Adaptor, AdapterHub does not provide support for more complex adaptation cases, such as using multiple objectives, scheduling, or extended evaluation. However, since both libraries build upon the HuggingFace Transformers library \cite{Wolf2019HuggingFacesTS}, their close integration is feasible.

If the robustness of models to heuristic shortcuts \cite{McCoy2019RightFT} is the primary goal, the Robustness~Gym library \cite{goel-etal-2021-robustness} provides a comprehensive evaluation over an extendable set of different kinds of heuristic biases.
Robustness~Gym provides much deeper evaluation compared to \Adaptor Evaluators, and could be integrated as an \Adaptor Evaluator.
Unlike Robustness Gym, \Adaptor enables an evaluation of robustness also on generative tasks, with specified out-of-domain data sets.

\section{\Adaptor Design}
\label{sec:framework_design}

\begin{figure*}
\centering
\includegraphics[width=\textwidth]{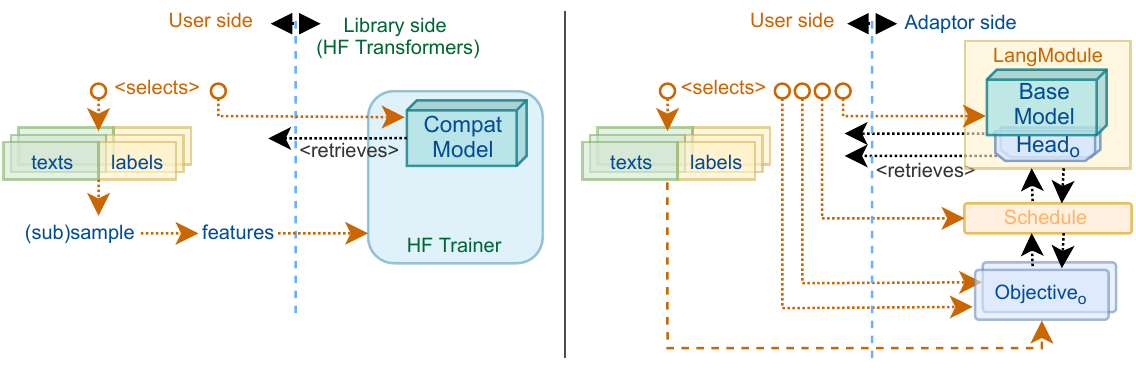}%
\vspace*{-.5\baselineskip}
\caption{A comparison of interaction with a model-centric HuggingFace Trainer (left) and objective-centric \Adaptor (right): While in model-centric approach, user resolves text processing, sampling and encoding compatible with selected model of specific objective, objective-centric approach delegates these functionalities to Objective instances.
Explicit definition of Objectives and Schedule on \Adaptor's user side makes otherwise complex multi-objective and custom-schedule experiments transparent and reproducible.}
\label{fig:user_flow}
\end{figure*}

This section describes the structure and functions of the \Adaptor framework.
We introduce its primary components bottom-up.
Figure~\ref{fig:user_flow} depicts the relations of these components and compares user interaction with the traditional model-centric pipeline.

\subsection{LangModule}
\label{sec:lang_module}


A LangModule instance provides a management of inputs, outputs and objective-specific model components, referred to as \textit{heads}. Once an objective with given LangModule is instantiated, an objective-compatible model is either initialised, or given by the user (see Section~\ref{sec:objective}) and the parameters of this model are merged with the parameters of the previously-registered objectives.

\iffalse
The merge works as follows: If no previous objective was registered, then the model of the given objective is considered a base model. The second- and later-registered objective models are then merged with the base model. First, pairs of PyTorch modules of the same name in the base and the new model are identified. If the dimensions and weights of these modules match, the respective module of the newly-added model is then replaced with a module of the base model.

In the case of pre-trained transformers, the weights of heads are initialized randomly, resulting in the registration of a different head for each objective and sharing the remaining parameters. Users can control which parameters (not) to merge by explicitly setting their respective weights as (non-)equal.

It is possible to use LangModule with any PyTorch module that uses a HuggingFace tokenizer, compatible with the given neural module. Therefore, this framework is also suitable for other models, such as recurrent networks.
\else
The merge works as follows: If no previous objective was registered, then the model of the given objective is considered a base model. The models of the second- and later-registered objectives are then merged with the base model: first, pairs of PyTorch modules of the same name in the base and the new model are identified. If the dimensions and weights of these modules match, the respective module of the newly-adding model is replaced with a module of the base model.

In the case of pre-trained transformers, the weights of heads are initialized randomly by default, resulting in a registration of a distinct head for each objective and sharing the remaining parameters. Users can control which parameters (not) to merge by explicitly setting their respective weights as (non-)equal.

It is possible to use LangModule with any PyTorch module that uses a HuggingFace tokenizer, compatible with the given neural module. Therefore, LangModule is also suitable for other models such as recurrent networks.
\fi

\subsection{Objective}
\label{sec:objective}

Objectives are the primary component of \Adaptor's training pipeline. Most importantly, an Objective serves two functions: sample encoding and loss computation. By implementing these and choosing the type of a model's head, \Adaptor users can define and experiment with novel training objectives. If they additionally provide an explicit definition of the Objective's model (the \mintinline{python}{objective_module} attribute), the new objective does not even have to comply with common model heads; shared parameters of the given \mintinline{python}{objective_module} would still be merged with the given \mintinline{python}{lang_module}.

If no \mintinline{python}{objective_module} is given, the Objective will request that a LangModule assigns the Objective a module of the Objective's default \mintinline{python}{compatible_head} (see~Section \ref{sec:lang_module}).

Additionally, every Objective instance performs its own logging, evaluation, and state updates, such as its convergence, based on a valuation of given  \mintinline{python}{val_evaluators}, or draws a progress bar, based on the state of its sample iteration. However, the training flow is guided by a Schedule (see Section~\ref{sec:schedule}). 
Objectives can implement custom data sampling, but if possible, we recommended to do so in a custom Schedule instance.

Since data encoding is also objective-specific, Objectives expose a higher-level user interface of data inputs than other frameworks: instead of encodings, users provide an Objective with a \mintinline{python}{texts_or_path} and a  \mintinline{python}{labels_or_path} containing raw texts and respective labels. \Adaptor provides an implementation of standard Objectives for sequence and token classification and sequence-to-sequence tasks. When implementing a custom Objective, note that sampling and encoding are performance bottlenecks on current high-end GPUs.

\subsection{Schedule}
\label{sec:schedule}

Schedules control the training flow through the interfaces provided by HuggingFace Transformers library. Primarily, they deliver 1)~a set of standard stopping strategies based on the state of the Objectives and 2)~an IterableDataset instance, constructed by sampling Objectives according to a sampling strategy implemented in its \mintinline{python}{_sample_objectives}. A Schedule also ensures that outputs of distinct \mintinline{python}{lang_modules}' heads are delivered to the respective Objectives for loss computation.

This relatively complex sampling framework provides a very simple interface for custom Schedule implementations (see Section~\ref{sec:data_sampling}). For instance, a pre-defined \mintinline{python}{ParallelSchedule} is implemented with three lines of code (see Figure~\ref{fig:schedule}).

\subsection{Adapter}
\label{sec:adapter}

An Adapter is customization of the HuggingFace Trainer with only minor adjustments.
Specifically, Adapter redirects loss computation to a Schedule, which further distributes outputs to corresponding Objectives and extends native training logs with logs of Objectives' Evaluators.
Furthermore, Adapter adjusts persistence of the models so that a model of every head can be reloaded without the use of \Adaptor, by simply using HuggingFace Transformers' \mintinline{python}{AutoModelForXY.from_pretrained}.

Based on the actively-developed HuggingFace Transformers library, the \Adaptor allows its users to benefit from all other native features of HuggingFace Transformers, such as the support for the most recent models, custom logging platforms, or distributed parallel training.
Furthermore, it can simplify integration with other custom libraries (see Section~\ref{sec:adapt_frameworks}).

\section{Experiments}
\label{sec:experiments}

\def\ob#1){\textit{\textbf{#1})}}
\begin{table*}
\tabcolsep1dd
\centerline{\scalebox{0.83}{%
\begin{tabular}{@{}l@{}l@{ \ }l@{ \ }ll@{ \ }l@{ }lll@{}}
\toprule
            & Schedule & \ \ \ Objectives               & BLEU$_{\textsc{ID}}$ & BLEU$_{\textsc{AD}}$ & BLEU$_{\textsc{OOD}}$ & BERTS$_{\textsc{ID}}$ & BERTS$_{\textsc{AD}}$ & BERTS$_{\textsc{OOD}}$ \\
\midrule
Pre-training &          & \ob1) Seq2Seq$_{\textsc{ID}}$              & 28.18      & \hphantom{0}5.34         & 0.91         & 0.833         & 0.738          & 0.671          \\ 
             & Sequent. & \ob2) Seq2Seq$_{\textsc{ID}}$+ BackTr$_{\textsc{AD}}$   & \hphantom{0}5.10 & 15.01  & 2.57  & 0.740  & 0.805   & 0.733   \\ 
             &          & \ob\llap*3) Seq2Seq$_{\textsc{ID}}$+ Seq2Seq$_{\textsc{AD}}$ & \hphantom{0}4.96 & 17.37  & 2.64  & 0.756  & 0.816   & 0.726   \\ 
             & Parallel & \ob4) Seq2Seq$_{\textsc{ID}}$+ BackTr$_{\textsc{AD}}$   & 31.06 & 16.99  & 2.46  & 0.852  & 0.817   & 0.722   \\ 
             &          & \ob\llap*5) Seq2Seq$_{\textsc{ID}}$+ Seq2Seq$_{\textsc{AD}}$ & 29.72 & 18.55  & 2.98  & 0.843  & 0.813   & 0.732   \\ 
\midrule
Fine-tuning  &          & \ob6) Seq2Seq$_{\textsc{ID}}$             & 37.97        & 17.62         & 6.50         & 0.875         & 0.808          & 0.758          \\ 
             &          & \ob7) BackTr$_{\textsc{AD}}$               & 30.34 & 22.98  & \llap11.08  & 0.869  & 0.834   & 0.799   \\ 
             & Parallel & \ob8) Seq2Seq$_{\textsc{ID}}$+ Denois$_{\textsc{AD}}$  & 38.96 & 13.37  & 6.87  & 0.876  & 0.782   & 0.761   \\ 
             &          & \ob9) Seq2Seq$_{\textsc{ID}}$+ BackTr$_{\textsc{AD}}$   & 38.25 & 21.47  & 9.03  & 0.873  & 0.831   & 0.791   \\ 
             &          & \ob\llap{*1}0) Seq2Seq$_{\textsc{ID}}$+ Seq2Seq$_{\textsc{AD}}$ & 40.72 & 23.35  & 6.97  & 0.880  & 0.836   & 0.772   \\  
\bottomrule
\end{tabular}%
}}
\caption{We evaluate the features of \Adaptor on multi-objective domain adaptation in machine translation: our experiments compare the BLEU score and BERTScore of unsupervised adaptation (\textit{Seq2seq} + \textit{Denoising} or \textit{Back-Translation}) applied in different schedules, to \textit{no} adaptation \ob(1, 6) and a hypothetical supervised adaptation \ob(*3, *5, *10).
Results show that the Parallel schedule eliminates \textit{catastrophic forgetting} and that unsupervised Back-translation is able to reach performance that is close to the supervised adaptation.}
\label{tab:results}
\end{table*}

We use \Adaptor in a set of domain adaptation experiments for a machine translation use-case, aiming to answer the following research question: \textbf{How well can unsupervised objective(s) substitute labeled parallel data}. 
In our methodology, we permute the easily-configurable parts of \Adaptor's training configuration\footnote{Our code is available on \url{https://github.com/gaussalgo/adaptor/tree/reprod/demo.py}} and compare the results of the resulting model to a baseline adaptation scenario.
We experiment with an architecture identical to the base model of~\citet{attention}, with a configuration of~\citet{mariannmt}.

\textbf{Data.}
We train the model on English-to-Czech translations on different domains of OPUS \cite{tiedemann-2012-parallel} chosen for their significant distinctiveness: we use \textit{Wikimedia} as a large-scale, supervised domain (denoted as \textit{in-domain}, i.e.\ ID), \textit{OpenSubtitles} as an Adapted Domain (AD) and \textit{Bible} for the evaluation of a model's robustness on Out-Of-Domain (OOD) samples.

\textbf{Pre-training vs.\ fine-tuning.}
We simulate two basic scenarios: training the model from a random initialization and fine-tuning the existing translation model with no control over its pre-training data.
In the latter cases, we perform fine-tuning from the checkpoint of~\citet{tiedemann-thottingal-2020-opus}.
\looseness=-1

\textbf{Schedules.} 
We implement and experiment with two objective schedules: i)~\textbf{Sequential} schedule, sampling and differentiating the model sequentially by each objective until convergence by evaluation loss, or for a maximum of 100,000 updates. 
ii)~\textbf{Parallel} schedule, concurrently sampling training batches uniformly from every given objective. 
Using gradient accumulation, we differentiate the model based on \textit{all} given objectives. 
We perform updates until the convergence of \textit{all} objectives, or for a maximum of 50,000 updates for each objective.

\textbf{Objectives selection.}
We implement and experiment with the following \Adaptor objectives:
\begin{itemize}
    \item \textbf{Sequence-to-sequence} (seq2seq) objective, as introduced by \citet{attention}, maps a combination of encoder inputs in the source language and previously-generated outputs as decoder inputs to a distribution over the next-predicted tokens.
    \item \textbf{Denoising} objective introduced by \citet{lewis2020bart} is an unsupervised instance of the seq2seq objective that performs random token permutation on the input and trains the model to map such `noisy` text to the original version of the input. We use this objective on the target-data domain to enhance its comprehension by the model.
    \item \textbf{Back-translation} objective, as used e.g.\ by \citet{sennrich-etal-2016-improving} is also an unsupervised seq2seq objective, which uses an external translator in reverse direction to obtain pseudo-inputs. This objective is profitable when we have unlabeled data of the target domain.
\end{itemize} 

Using these components, we construct the following experiments: 
\begin{itemize}
    \item \textbf{Baselines:} pre-training \ob(1)~and fine-tuning \ob(6) on ID data from a domain different from the Application Domain (AD) using a single traditional \textit{seq2seq} objective.
    \item \textbf{Sequential adaptation:} we pre-train using \textit{seq2seq} on ID and afterwards adapt using either unsupervised \textit{Back-translation} \ob(2), or supervised \textit{seq2seq} \ob(3) on AD to quantify the unsupervised adaptation gap.
    \item \textbf{Parallel adaptation:} we concurrently train on both \textit{seq2seq} and another unsupervised objective: \textit{Back-translation} \ob(4, 9) and \textit{Denoising} \ob(8). Again, we compare the gap to the supervised situation \ob(5, 10).
\end{itemize}

\subsection{Results}
\label{sec:results}

Table~\ref{tab:results} evaluates the base transformer after the given number of updates on held-out deduplicated validation splits of In-Domain (ID), Adapted-Domain (AD), and the third Out-Of-Domain (OOD) data. Note that the results for the BLEU score are properly comparable only within the same domain.

We observe that the model trained on a single domain \ob(1, 6) degrades on \textit{all} other domains.
In a pre-training scenario, domain robustness improves when incorporating data of adapted domain in \textit{any} objective.
However, in a sequential schedule, we observe catastrophic forgetting towards any most-recent domain of adaptation \ob(2, 3).
This is improved by using the Parallel schedule for a negligible price of in-domain accuracy \ob(4, 5).

In the fine-tuning scenario, we show that incorporating unsupervised Back-translation to AD \ob(7, 9) improves ID BLEU comparably to supervised adaptation \ob(10). Interestingly, Denoising on AD \ob(8) improves in-domain performance but seems less efficient than Back-translation.

\subsection{\Adaptor Usage Complexity}
To give an idea about the relative complexity of using \Adaptor as compared to model-centric frameworks, we compare selected measurable code features of the complexity of our experimental implementation to an example implementation using the HuggingFace Trainer\footnote{For reference, we use \textit{run\_translation.py} example script on HuggingFace Transformers GitHub, version 4.17.0.}.
We pick the experiment of supervised pre-training + unsupervised fine-tuning, including evaluation, in the sequential schedule \ob(2), as this can still be addressed using HuggingFace Transformers relatively easily;
Implementing the parallel multi-objective schedule in the Transformers framework would require major customisations of selected model and Trainer objects.

The training script using HuggingFace Trainer contains 654 lines of code, 135 variable assignments, 186 method calls and the initialisation of 9~custom objects.
Additionally, in the pre-training + fine-tuning framework, this script has to be run twice, initialising the second training from the selected checkpoint of the first one, with updated configurations.
Back-translated pseudo-labels are generated by a different script, not included in this assessment.

Using \Adaptor, we construct an equivalent routine from the provided demo script. Our implementation contains 124 lines of code, 31~variable assignments, 37~method calls and the initialisation of 14~custom objects. 
Despite its brevity, our script wraps the whole training process, and hence, together with the associated version of \Adaptor or its fork, it provides a reproducible fingerprint of the experiment.

\section{Conclusion and Future Work}
\label{sec:conclusion}

This paper introduces the \Adaptor library, which provides objective-centric training framework well-suitable for multi-task and multi-domain training scenarios, and the development of novel objectives and sampling schedules.
We find that even in the conventional single-objective training routines, \Adaptor can reduce volumes of custom implementation and increases readability and reproducibility.
Having used \Adaptor already for several production use cases, we are happy to share it with the NLP community.

Our future work aims to further enhance \Adaptor's user comfort with existing and novel unsupervised objectives, dynamic schedules, and demonstrations on novel use cases.

\section{Broader Impact}
\label{sec:broader_impact}

Thanks to the ubiquity of objective-centric training, \Adaptor can accelerate the applicability of the most recent research in multi-task and multilingual modeling and enrich the research with the practical experience of the industry.

We further identify the benefits of \Adaptor's definite training pipelines in saving unnecessary financial and environmental expenses of reproducing the reported results of large language models, otherwise often including expensive hyperparameter optimization over unreported parameters. 
Due to these aspects, \Adaptor could also ease the spread of state-of-the-art language technologies to under-resourced languages and more specialized domains with a sufficient amount of unsupervised sources.

Finally, objective-centric training might help expose the potential of unsupervised objectives to the generalization and interpretability of models.
\Adaptor can foster the research in unsupervised learning by lowering the relatively high entry level of technical proficiency needed for experimentation with novel language objectives.
%

\bibliography{stefanik}
\bibliographystyle{acl_natbib}

\end{document}